\documentclass[conference]{IEEEtran}
\IEEEoverridecommandlockouts
\usepackage{subfigure}
\usepackage{cite}
\usepackage{amsmath,amssymb,amsfonts}
\usepackage{algorithmic}
\usepackage{textcomp}
\usepackage{caption}
\usepackage{graphicx}
\usepackage{multirow}

\usepackage[table,xcdraw]{xcolor}

\def\BibTeX{{\rm B\kern-.05em{\sc i\kern-.025em b}\kern-.08em
    T\kern-.1667em\lower.7ex\hbox{E}\kern-.125emX}}
\usepackage{comment}
\usepackage{authblk}

\usepackage{amssymb}
\usepackage{fancyhdr}
\fancypagestyle{firstpage}
{
    \fancyhead[L]{\footnotesize © 2023 IEEE.  Personal use of this material is permitted.  Permission from IEEE must be obtained for all other uses, in any current or future media, including reprinting/republishing this material for advertising or promotional purposes, creating new collective works, for resale or redistribution to servers or lists, or reuse of any copyrighted component of this work in other works. This paper is accepted at the 24th International Symposium on Quality Electronic Design (ISQED) 2023.}
    \fancyhead[R]{}
}

\begin{document}

\title{DeepAxe: A Framework for Exploration of Approximation and Reliability Trade-offs in DNN Accelerators}

\author{
{Mahdi Taheri}\textsuperscript{1}\textsuperscript{*}, \
{Mohammad Riazati}\textsuperscript{2}\textsuperscript{*},\
{Mohammad Hasan Ahmadilivani}\textsuperscript{1},\\
{Maksim Jenihhin}\textsuperscript{1},\
{Masoud Daneshtalab}\textsuperscript{1,2},\
{Jaan Raik}}
\author[2]{Mikael Sjödin}
\author[2]{Björn Lisper}
\affil[1]{Tallinn University of Technology, Tallinn, Estonia}
\affil[2]{Mälardalen University, Västerås, Sweden}
\affil[1]{mahdi.taheri@taltech.ee}

\maketitle

\thispagestyle{firstpage}

\begin{abstract}
While the role of Deep Neural Networks (DNNs) in a wide range of safety-critical applications is expanding, emerging DNNs experience massive growth in terms of computation power.  
It raises the necessity of improving the reliability of DNN accelerators yet reducing the computational burden on the hardware platforms, i.e. reducing the energy consumption and execution time as well as increasing the efficiency of DNN accelerators.
Therefore, the trade-off between hardware performance, i.e. area, power and delay, and the reliability of the DNN accelerator implementation becomes critical and requires tools for analysis.

In this paper, we propose a framework DeepAxe for design space exploration for FPGA-based implementation of DNNs by considering the trilateral impact of applying functional approximation on accuracy, reliability and hardware performance. 
The framework enables selective approximation of reliability-critical DNNs, providing a set of Pareto-optimal DNN implementation design space points for the target resource utilization requirements.
The design flow starts with a pre-trained network in Keras, uses an innovative high-level synthesis environment DeepHLS and results in a set of Pareto-optimal design space points as a guide for the designer. The framework is demonstrated on a case study of custom and state-of-the-art DNNs and datasets.

\end{abstract}
\let\thefootnote\relax\footnote{* These authors contributed equally}

\section{Introduction}
In the past decades, Deep Neural Networks (DNNs) demonstrated a significant improvement in accuracy by adopting intense parameterized models \cite{taheri2022dnn}.
As a consequence, the size of these models has drastically increased imposing challenges in deploying them on resource-constrained platforms \cite{gholami2021survey}. 
FPGAs are a widely used solution for flexible and efficient DNN accelerator implementations and have shown superior hardware performance in terms of latency and power \cite{riazati2022autodeephls}. 


In practice, deployment of a DNN accelerator for the safety-and mission-critical applications  (e.g., autonomous driving) requires addressing the trade-off between different design parameters of \textit{hardware performance}, e.g., area, power, delay, and \textit{reliability}.
A compromise between conflicting requirements can be achieved by simplifying the implementation to sacrifice the precision of results but benefiting from lower resource utilization, energy consumption, and higher system efficiency. \textit{Approximation Computing (AxC)} is one of such concepts in hardware design \cite{choudhary2022approximate}.

Moreover, the assessment of the reliability of DNN accelerators is a challenging issue by itself. Reliability of DNNs concerns DNN accelerators' ability to execute correctly in the presence of faults \cite{ibrahim2020soft} originating from either the environment (e.g., soft errors, electromagnetic effects, temperature variations) or from inside of the chip (e.g., manufacturing defects, process variations, aging effects) \cite{shafique2020robust}. 
The ability to tolerate the impact of faults on the output accuracy is called \textit{fault resiliency} and, in practice, it is one of the contributors to the DNN accelerators' reliability  \cite{booth2022algorithm}. 
DNNs are known to be inherently fault-resilient due to the high number of learning process iterations and also several parallel neurons with multiple computation units.
Nevertheless, faults may impact the output accuracy of DNNs drastically \cite{bosio2019reliability}, and in case of resource-constrained critical applications, DNNs' fault resiliency is required to be evaluated and guaranteed \cite{cavagnero2022fault} 
\cite{siddique2021exploring}.

The complexity of such evaluation motivates an \textit{automated tool-chain} with AxC and resiliency analysis to support \textit{Design Space Exploration (DSE)} for DNN accelerators already at the early design stage, i.e. starting from a high-level description.

High-Level Synthesis (HLS) tools bridge high-level programming and hardware implementation and allow overcoming the complexity of the process and reducing the design time. Recently, DNN-tailored HLS tools were proposed, e.g., CNN2gate\cite{2}, fpgaConvNet \cite{2} and DeepHLS \cite{9294881}. Such tools are capable of providing a synthesizable C implementation of DNNs for FPGAs from a high-level description in a language such as e.g., Keras. 

This paper presents a novel framework and a fully automated tool-chain DeepAxe to provide a design space exploration for FPGA-based implementation of DNN accelerators by analyzing
approximation and soft-error reliability trade-offs. 
To the best of our knowledge, this is the first framework that holistically considers both the transient fault resiliency and hardware performance of DNN accelerators as design parameters. DeepAxe is empowered by  techniques for quantizing the networks and providing the capability of substituting the exact computing (ExC) units of the network with AxC units and identifying the optimal design points for selective approximation. 
 
DeepAxe uses the Keras description of a DNN as the input and is capable of providing an FPGA-ready approximated and transient-fault-resilient inference implementation of the network based on the design parameters selected based on the DSE results. The main contributions in this work are as follows:
\begin{itemize}
    \item A methodology for selective approximation of reliability-critical DNNs providing a set of Pareto-optimal DNN implementation design space points for the target resource utilization requirements. 
    \item A framework DeepAxe for holistic exploration of approximation and reliability trade-offs in DNN accelerator FPGA-based implementation that enables assessing the trilateral impact of approximation on accuracy, reliability, and hardware performance.
    \item Integration of the fully automated DeepAxe tool-chain into the DeepHLS environment. 
    \item Demonstration and validation of the framework on representative custom and state-of-the-art DNNs and datasets. 
    
\end{itemize}

The rest of the paper is organized as follows. Related works are discussed in Section II, the DeepAxe methodology and framework are presented in Section III, the experimental setup and results are provided in Section IV, and finally, the work is concluded in Section V.

\section{Related Works}
The advantages of implementing and deploying DNNs on FPGAs are advocated in several recent works. The existing FPGA-based tool-chains to map Convolutional Neural Networks (CNNs) are presented in the surveys \cite{venieris2018toolflows, guo2019dl, abdelouahab2018accelerating, molina2022high}. The FINN framework \cite{FINN} is released by Xilinx  for the exploration of quantized CNNs' inference on FPGAs that also provides  customized data-flow architectures for each network. 
Research works \cite{riazati2022autodeephls} and \cite{1} provide Register-Transfer Level (RTL) models using conventional synthesis tools, e.g., Vivado HLS, where the outputs can be directly synthesized on an FPGA. Heterogeneous systems are also another design strategy in the automated tool-chains that propose hardware-software co-design \cite{1, ghaffari2020cnn2gate, mousouliotis2020cnn}. In these designs, computational units, e.g., addition, or multiplication, are mainly implemented on Processing Logic (PL) that is controlled by a control unit in a CPU using a dedicated framework, e.g., OpenCL \cite{stone2010opencl}.
 
Using Fixed-point (FxP) data type instead of Floating Point (FP) is becoming more popular due to the lesser resource utilization while keeping the output accuracy degradation at an acceptable level \cite{1,guo2016angel, sharma2016high}. Throughout the literature, comprehensive simulations exist that prove that merely an 8-bit data type for MAC operations in DNN execution is sufficient to provide a practical accuracy along with favorable resource utilization \cite{jouppi2017datacenter, sarwar2018energy}. In this work, we considered 8-bit as the base data type for the simulations and implementations.

A number of works in the literature explore the reliability of the DNNs \cite{hong2019terminal, kundu2020high}. Some works examine the impact of different fault models on the basis of a number of layers in DNNs and different data types \cite{li2017understanding}. Studying the significant impact of transient faults vs permanent faults is also done by \cite{zhang2018analyzing}. The fault analysis of exact DNNs has drawn a lot of attention in the state-of-the-art research \cite{deepvigor}, and only recently, researchers have started to investigate also the reliability of approximated DNN accelerators (AxDNNs) \cite{siddique2021exploring}. A somewhat expected conclusion in \cite{kundu2020high} is that the error induced by approximation, along with the faults in the DNN structure, are not evenly propagated. The impact of a fault may differ based on different parameters, like fault type, fault location, the approximation error resiliency for each layer, etc. To the best of our knowledge, none of these works explored the impact of using different combinations of approximated layers of a DNN in the presence of transient faults on the reliability, accuracy and delay/resource utilization of the target DNN accelerator.

The approach proposed in this paper goes beyond the state of the art by establishing a fully automated tool for enabling efficient AxC in FPGA-based DNN accelerators aimed at reliability-critical applications. The proposed DeepAxe framework is integrated into DeepHLS environment \cite{9294881}, which is capable of providing completely synthesizable code for efficient FPGA implementations. In particular, this work extends DeepHLS with fault simulation, resiliency analysis and also the use of AxC. The new features allow providing the designers a guideline to choose optimal configurations based on specific requirements for latency, accuracy, resource utilization, and fault resiliency.

\section{Proposed Methodology}

 \begin{figure}[h]
    \centering
    \includegraphics[width = 0.47\textwidth]{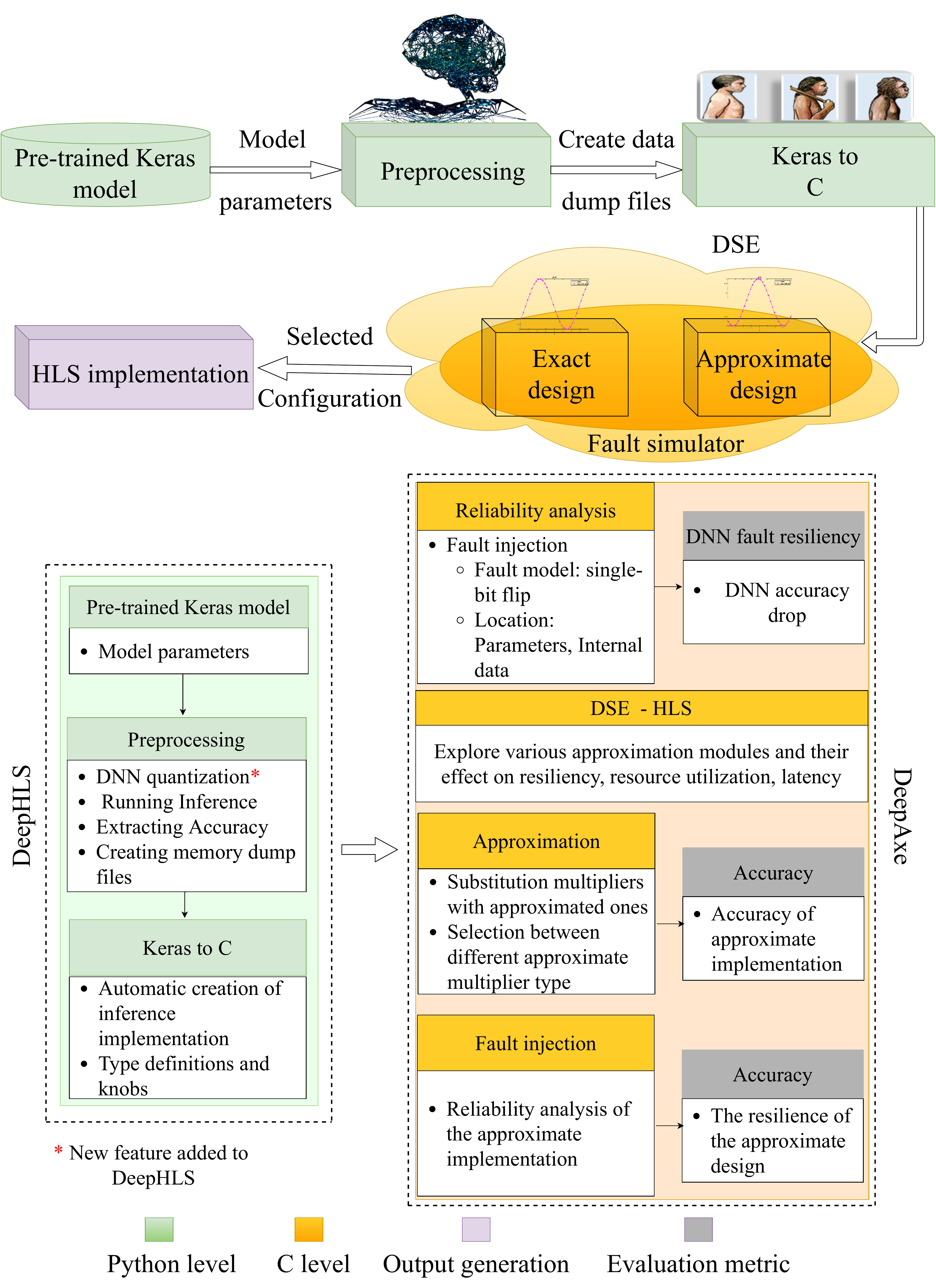}
    \caption{DeepAxe methodology flow}
    \label{method}
\end{figure}

\begin{figure}[h]
    \centering
    \includegraphics[width = 0.35\textwidth]{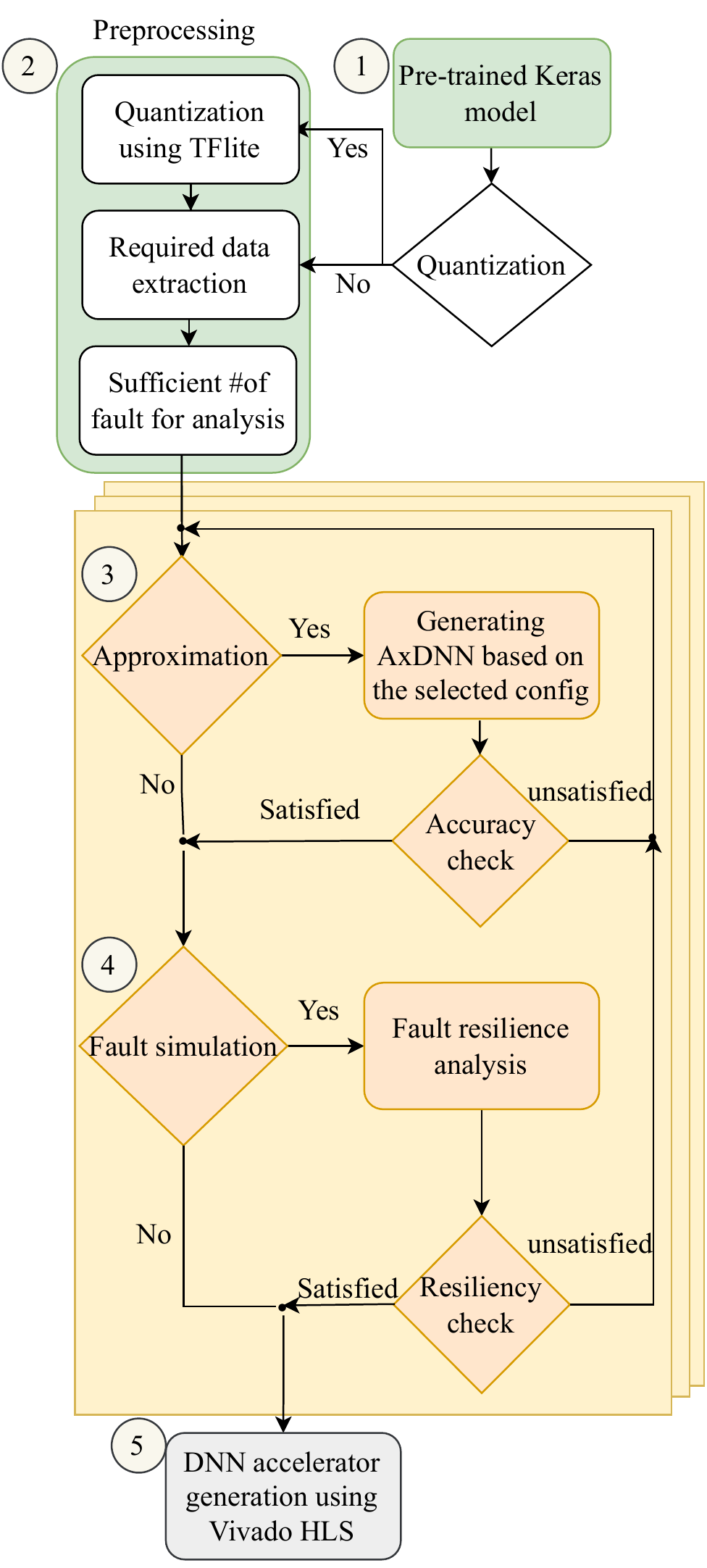}
    \caption{DeepAxe flowchart}
    \label{flowchart}
\end{figure}

Fig. \ref{method} illustrates the methodology flow established in the DeepAxe tool-chain for reliability and hardware performance analysis of approximated DNN hardware accelerators. DeepAxe is a framework taking the DNNs' \emph{Pre-trained Keras model} description as the input. Then, DeepAxe feeds the extracted model parameters through the flow to apply the initialization needed before creating the C code. The design, training and test of the DNNs are performed in Python, the \emph{Preprocessing} step is seamlessly integrated into the same environment and is responsible for extracting the required data for the next step. 

DeepAxe also supports quantizing the network down to 8-bit INT as a part of the preprocessing step. For this purpose, a full quantization is implemented, targeting all activations, weights and biases. The framework first takes the description of the network in Keras, and then uses the TFlite library to generate a training-aware quantized network. The user can replace their preferred Keras-based quantization library to the tool-chain for this step. The main output of this step is the quantized network's parameters (i.e., weight/bias) and also the files containing the memory dump of the test data. Specifically, the \emph{Keras to C} step implies converting all the above-mentioned parameters to multidimensional arrays in C format. The output accuracy of the generated network is also provided at this step and is kept as a baseline for the further steps of the methodology. 

\emph{Reliability analysis} relies on a fault injection (FI) in C, assuming the single bit-flip faults in the network's activation layers for resiliency assessment. While the multiple-bit fault model is more accurate, it requires a prohibitively large number of fault combinations to be considered ($3^n - 1$ combinations, where $n$ is the number of bits). Fortunately, it has been shown that high fault coverage obtained using the single-bit model results in a high fault coverage of multiple-bit faults \cite{bushnell2004essentials}. Therefore, a vast majority of practical FI and test methods are based on the single-bit fault assumption.

The reliability analysis step applies the accuracy drop comparison of the network-under-test as the assessment metric.
\emph{Approximate design} (see the yellow region in Fig. \ref{method}) refers to the selective approximation of DNNs by layers provided by DeepAxe. It instruments the user with the flexibility of choosing between a) different AxC models provided by any library of approximate computing units, such as AxC multipliers in EvoApproxLib, and b) the subset of layers, for setting up different configurations of the network. As an example, in a network with $n$ computing layers (containing both convolutional and fully connected layers), the user has $2\textsuperscript{n}$ combinations for exploring the exact and approximate implementations for each layer individually. 

After choosing the preferred approximation configuration, the designer can go through the fault injector provided for the resiliency evaluation of the AxDNN. Eventually, the final design can be fed to the \emph{HLS implementation} step for DNN hardware accelerator generation process by the HLS tool.

To illustrate the DeepAxe methodology, the flowchart provided in Fig. \ref{flowchart} shows the step-by-step process from the beginning to the end of DeepAxe tool-chain. After providing the Keras description of the network in Step 1, the user can decide if they need to quantize the network. Then, the preprocessing step can be performed, enabling the user to apply a pre-analysis on the network to extract a sufficient number of faults for the reliability assessment, considering the number of its neurons. 

Steps 3 and 4 in Fig. \ref{flowchart} show an iterative process to examine different approximated DNN combinations and, accordingly, their fault resiliency analysis to build the DSE. By enabling the fault simulation process in Step 4, the user can follow the impact of their chosen AxC model and also the approximation configuration on the resiliency of the network compared to the other AxC model/configurations and also to the exact model. Finally, the selected design and its configuration are fed into the HLS tool for implementation.

It is noteworthy that all steps in the yellow box of Fig. \ref{method} can be iterative, and the user can repeat these steps to find the optimal point based on their requirements. For instance, the user might decide to analyze an assumed approximation configuration, i.e. AxC model for the multiplier and also the layers to approximate. If, after applying approximation, the accuracy check does not satisfy the user, they can try another approximation configuration. Once the requirements are satisfied, it is possible to proceed to the fault vulnerability analysis. If, after applying the fault injection, the resiliency of the network is also satisfying, the next step is generating the DNN accelerator based on the selected configuration.

\section{Experimental Results}

\subsection{Experimental Setup}

First, all DNNs are implemented, trained and tested in Keras. The required data for further steps of DeepAxe are also generated in the same environment. In the DeepAxe flowchart (Fig. \ref{flowchart}), the green parts, including steps 1 and 2, refer to the steps of the framework implemented in this high-level environment.
Both a three-layer MLP and LeNet-5, trained on the MNIST dataset, and AlexNet, trained on the CIFAR-10 dataset, are representative DNNs and efficient to perform the validation of the proposed methodology and framework. 
All networks use ReLu as an activation function. 

All networks are quantized down to 8-bit INT data type, including all activations, weights, and biases, by using the TFlite \cite{david2021tensorflow} library in Python.
The yellow parts in Fig. \ref{flowchart} are implemented in C. Simulations are performed on 2 x Intel Xeon Gold 6148 2.40 GHz (40 cores, 80 threads per node) with 96GB RAM. To speed up the simulation process, DeepAxe supports multi-thread parallelism, and users can benefit from this feature based on the number of cores their CPU provides. 

All implementations in C are synthesizable by DeepHLS. The approximate multipliers in the C implementation of the network (referring to step 3 in Fig. \ref{flowchart}) are adopted from the C codes provided by EvoApproxLib library \cite{evoapprox16}. In this paper, three 8-bit INT approximate multipliers are picked from EvoApproxLib with different error, area, and power characteristics reported in Table \ref{mult}. The error parameters reported in this table are as follows:
\begin{itemize}
    \item MAE - Mean Absolute Error (Mean Error Magnitude)
    \item WCE - Worst-Case Absolute Error (Error Magnitude / Error Significance)
    \item MRE - Mean Relative Error (Mean Relative Error Distance)
    \item EP - Error Probability (Error Rate)
\end{itemize}
Power (power consumption in $mW$) and area (area on the chip in $\mu m^2$) are also reported as the design parameters in the last two columns of the table. 
To show the hardware characteristics of the output AxDNN, the Lookup Table (LUT) and Flip Flop (FF) utilization, as well as the number of required clock cycles for a one-time execution of the output AxDNN accelerator, are reported as the results based on the reports produced by Xilinx Vivado HLS tool on a Xilinx Spartan-7 FPGA with part number xc7s100-fgga676-1 and 100 MHz frequency.

\subsection{Fault simulator}
The fault simulator that is used in step 4 in Fig. \ref{flowchart} is implemented in the automated tool-flow of DeepAxe in a way that users can select the sufficient number of faults they need for their resiliency analysis. 
AxDNNs generated by step 3 in Fig. \ref{flowchart} are validated by means of fault injection over the test set. 

\begin{table}
\caption{Exact and approximate multipliers used in this paper and their parameters}
\label{mult}
\resizebox{0.5\textwidth}{!}{%
\begin{tabular}{|c|c|c|c|l|l|l|}
\hline
Circuit name     & MAE   & WCE  & MRE  & EP    & Power & Area  \\ \hline
Exact multiplier & 0.00  & 0.00 & 0.00 & 0.00  & 0.425 & 729.8 \\ \hline
mul8s\_1KVP      & 0.051 & 0.21 & 2.73 & 74.80 & 0.363 & 635.0 \\ \hline
mul8s\_1KV9 & \multicolumn{1}{l|}{0.0064} & \multicolumn{1}{l|}{0.026}  & \multicolumn{1}{l|}{0.90} & 68.75 & 0.410 & 685.2 \\ \hline
mul8s\_1KV8 & \multicolumn{1}{l|}{0.0018} & \multicolumn{1}{l|}{0.0076} & \multicolumn{1}{l|}{0.28} & 50.00 & 0.422 & 711.0 \\ \hline
\end{tabular}%
}
\end{table}

\emph{Random Fault Injection.} 
According to the adopted fault model, a random single bit-flip is injected into a random neuron in a random layer of the network, and the whole test set is fed to the network to obtain the accuracy of the network. This process is repeated several times to reach an acceptable confidence level which depends on the number of neurons and data representation bit length based on \cite{leveugle2009statistical}. 

To find the required number of repetitions for the fault simulation experiments, \cite{leveugle2009statistical} provides an equation to reach 95\% confidence level and 1\% error margin. However, it can pessimistically obtain a larger number, and the execution time of the iterative fault simulation experiments would be very long. Therefore, we have performed a fault simulation for each neural network to find a smaller number of experiments in a way that the difference of the average accuracy is less than 0.1\% in comparison with the average accuracy of the network achieved using the statistical fault injection approach \cite{leveugle2009statistical}. As a result, we have selected for injection 600, 800, and 1000 random single bit-flip faults for 3-layer MLP, LeNet-5, and AlexNet fault simulation, respectively.

\subsection{Validation Results}

\begin{figure*}[h]
    \centering
    \subfigure{\includegraphics[width=0.5\textwidth]{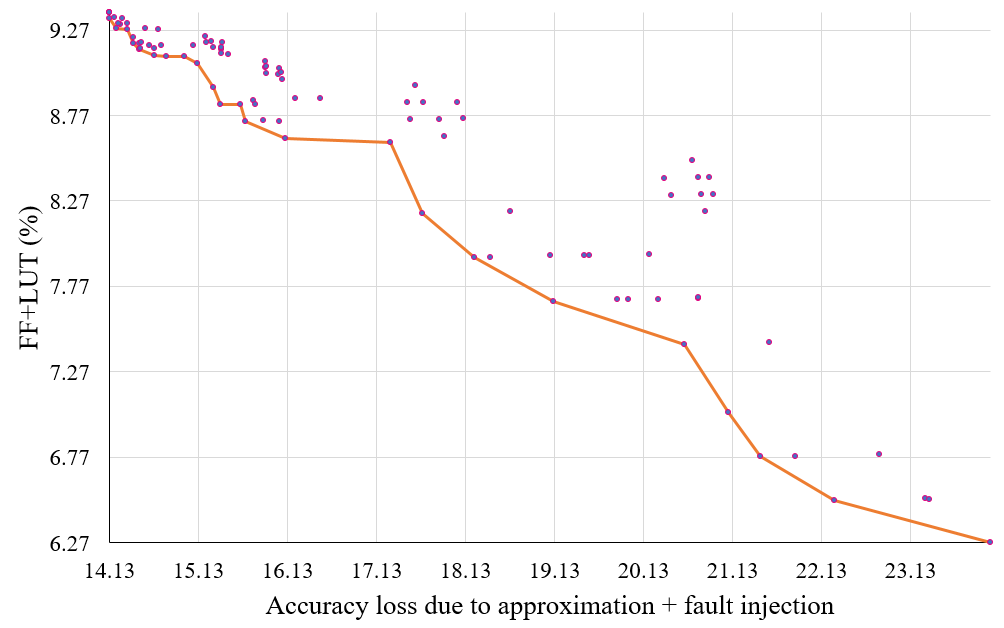}} 
    \subfigure{\includegraphics[width=0.49\textwidth]{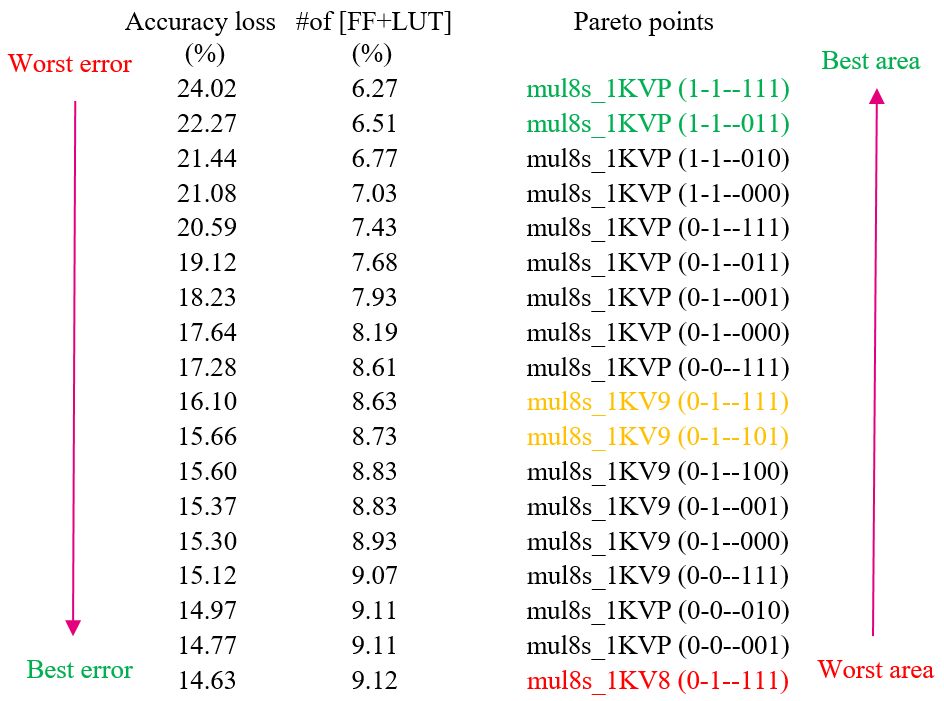}} 
    \caption{ (a) Resource utilization of the approximate implementation vs. accuracy drop when the approximate implementation is fault-simulated (b) Approximation configuration of each point on the  Pareto frontier}
    \label{pareto}
\end{figure*} 

\begin{figure*}[h!]
    \centering
    \subfigure{\includegraphics[width=0.69\textwidth]{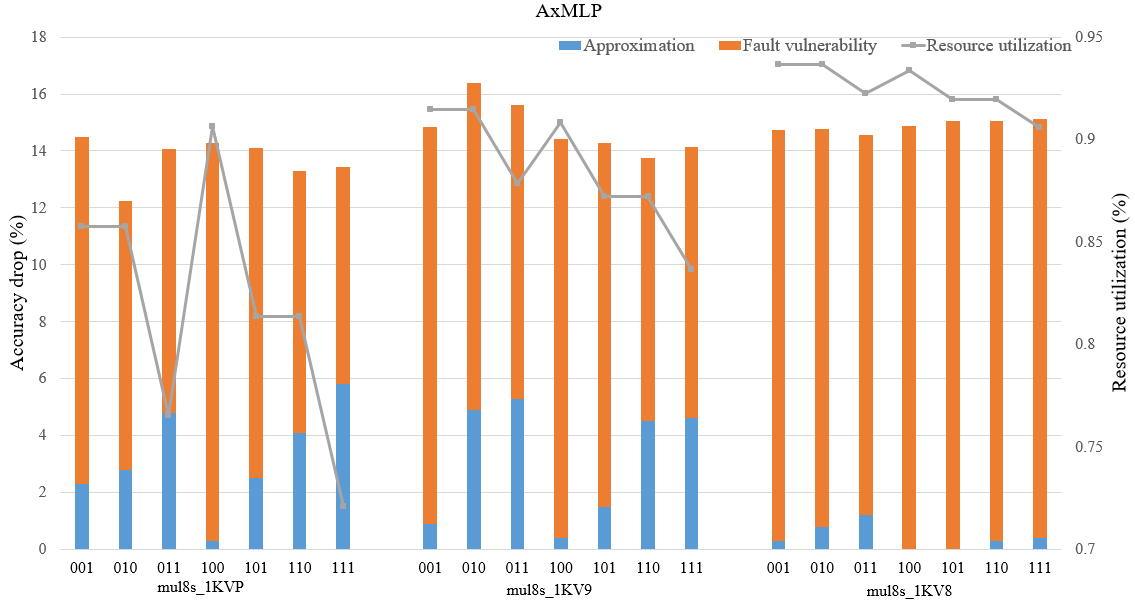}} 
    \subfigure{\includegraphics[width=0.69\textwidth]{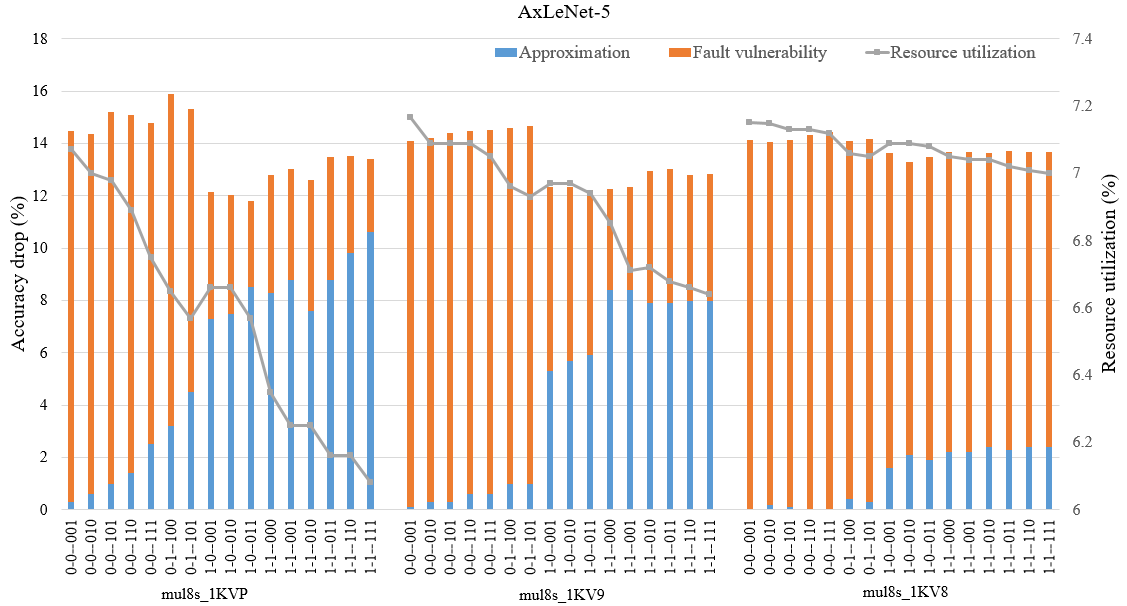}} 
    \subfigure{\includegraphics[width=0.69\textwidth]{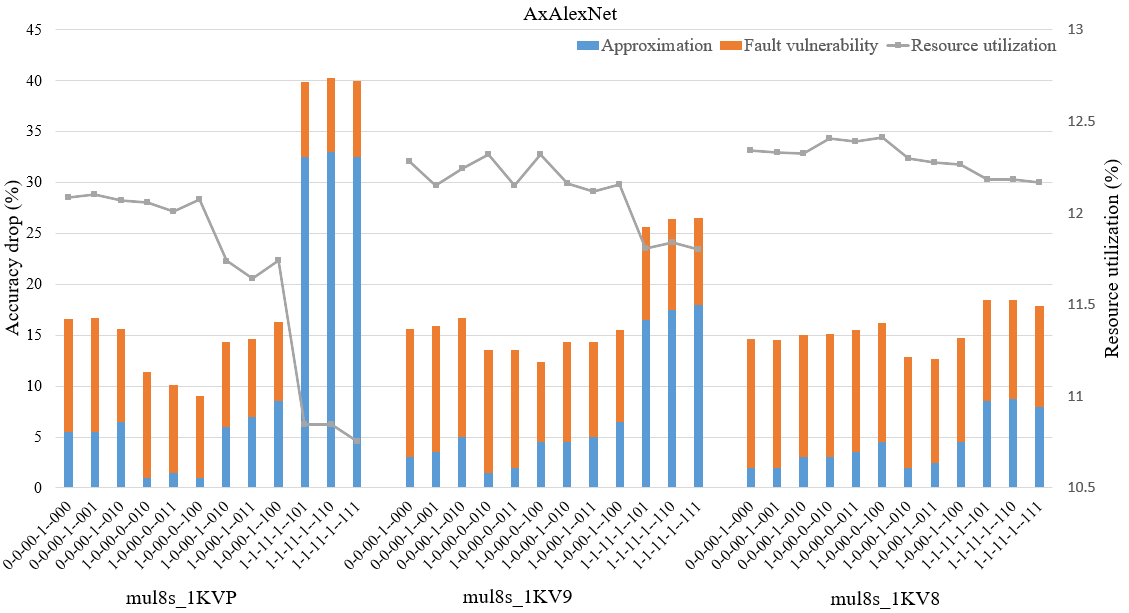}}     
    \caption{Reports of accuracy drop (due to approximation for different configurations), fault vulnerability, and resource utilization of (a) 3-layer MLP network, (b) LeNet-5 and (c) AlexNet}
    \label{ax-accuracy}
\end{figure*}

\begin{table}
\centering
\caption{Networks trained and quantized down to 8-bit INT for evaluation of this work}
\label{net}
\begin{tabular}{|c|c|c|c|}
\hline
Network & Dataset & \begin{tabular}[c]{@{}c@{}}Accuracy\\ 8-bit quantized network\end{tabular} \\ \hline
3-layer MLP & MNIST   & 80.40\%  \\ \hline
LeNet-5     & MNIST   & 85.80\%  \\ \hline
AlexNet     & CIFAR-10 & 78.50\% \\ \hline
\end{tabular}
\end{table}

\begin{table*}[h]
\centering
\caption{The impact of approximation configuration and fault injection for MLP, LeNet-5, and AlexNet.}
\label{all-results}
\resizebox{17.5cm}{!}{
\begin{tabular}{|c|c|c|c|c|c|c|l|}
\hline
\begin{tabular}[c]{@{}c@{}}DNN\\ dataset\end{tabular} &
  Multiplier &
  Layer configuration &
  \begin{tabular}[c]{@{}c@{}}Base \\ accuracy (\%)\end{tabular}  &
  \begin{tabular}[c]{@{}c@{}}Accuracy drop (\%)\\ $[$Exact network - \\ AxDNN$]$\end{tabular} &
  \begin{tabular}[c]{@{}c@{}}AxDNN accuracy drop (\%)\\ $[$AxDNN  - \\ FI on AxDNN$]$\end{tabular} &
  \begin{tabular}[c]{@{}c@{}}Latency\\ (\#of clk cycles)\end{tabular} &
  \multicolumn{1}{c|}{\begin{tabular}[c]{@{}c@{}}Resource utilization (\%)\\ \#of[FF + LUT] / \\ Total \#of[FF + LUT]\end{tabular}} \\ \hline
\multirow{5}{*}{\begin{tabular}[c]{@{}c@{}}MLP\\ MNIST\end{tabular}} &
  mul8s\_1KVP &
  111 &
   &
  5.8 &
  7.62 &
  206644 &
  \multicolumn{1}{c|}{0.72} \\ \cline{2-3} \cline{5-8}
 &
  mul8s\_1KVP &
  101 &
   &
  2.5 &
  11.62 &
  272180 &
  \multicolumn{1}{c|}{0.81} \\ \cline{2-3} \cline{5-8}
 &
  mul8s\_1KV9 &
  101 &
  80.40 &
  1.5 &
  12.78 &
  274740 &
  \multicolumn{1}{c|}{0.87} \\ \cline{2-3} \cline{5-8} 
 &
  mul8s\_1KV9 &
  100 &
   &
  0.4 &
  14.03 &
  274740 &
  \multicolumn{1}{c|}{0.90} \\ \cline{2-3} \cline{5-8}
 &
  mul8s\_1KV8 &
  001 &
   &
  0.3 &
  14.72 &
  285010 &
  \multicolumn{1}{c|}{0.95} \\ \hline
\multirow{5}{*}{\begin{tabular}[c]{@{}c@{}}LeNet-5\\ MNIST\end{tabular}} &
  mul8s\_1KVP &
  1-1--111 &
   &
  10.6 &
  2.82 &
  164864 &
  \multicolumn{1}{c|}{6.27} \\ \cline{2-3} \cline{5-8} 
 &
  mul8s\_1KVP &
  1-1--011 &
   &
  8.8 &
  4.67 &
  195584 &
  \multicolumn{1}{c|}{6.51} \\ \cline{2-3} \cline{5-8} 
 &
  mul8s\_1KV9 &
  0-1--111 &
  85.80 &
  1.7 &
  12.70 &
  206408 &
  \multicolumn{1}{c|}{7.93} \\ \cline{2-3} \cline{5-8} 
 &
  mul8s\_1KV9 &
  0-1--101 &
   &
  1.0 &
  13.66 &
  206504 &
  \multicolumn{1}{c|}{8.19} \\ \cline{2-3} \cline{5-8} 
 &
  mul8s\_1KV8 &
  0-1--111 &
   &
  0.7 &
  13.23 &
  175784 &
  \multicolumn{1}{c|}{9.12} \\ \hline
\multirow{5}{*}{\begin{tabular}[c]{@{}c@{}}AlexNet\\ CIFAR-10\end{tabular}} &
  mul8s\_1KVP &
  0-0-11-0--011 &
   &
  16.0 &
  9.12 &
  19933514
  &
  \multicolumn{1}{c|}{11.75} \\ \cline{2-3} \cline{5-8} 
 &
 mul8s\_1KVP &
  0-0-11-0--100 &
   &
  17.0 &
 10.41 &
20324170 &
 \multicolumn{1}{c|}{11.84}  \\ \cline{2-3} \cline{5-8} 
 &
  mul8s\_1KVP &
  0-0-00-0--001 &
   &
  2.0 &
  11.10 &
  20467530
 &
   \multicolumn{1}{c|}{12.35} \\ \cline{2-3} \cline{5-8} 
 &
  mul8s\_1KV9 &
  0-1-11-1--111 &
   &
  18.5 &
  9.58 &
  19799882
 &

 \multicolumn{1}{c|}{11.04} \\ \cline{2-3} \cline{5-8} 
 &
  mul8s\_1KV9 &
  0-1-11-1--110 &
   &
  17.5 &
  11.80 &
  19945802
 &
   \multicolumn{1}{c|}{11.93}  \\ \cline{2-3} \cline{5-8} 
 &
  mul8s\_1KV9 &
  0-0-00-0--001 &
   &
  3.0 &
  12.60 &
  20470090
 &
   \multicolumn{1}{c|}{12.45}   \\ \cline{2-3} \cline{5-8} 
 &
   mul8s\_1KV8 &
  1-1-11-1--110 &
  78.50 &
  6.5 &
  10.90 &
  20470090

 &
   \multicolumn{1}{c|}{12.18}   \\ \cline{2-3} \cline{5-8} 
 &
    mul8s\_1KV8 &
  0-1-11-1--111 &
   &
  6.0 &
  11.70 &
  20470090 &
   \multicolumn{1}{c|}{12.19}   \\ \cline{2-3} \cline{5-8} 
 &
  mul8s\_1KV8 &
  0-1-11-1--110 &
   &
  4.5 &
  12.00 &
  20470090
   &
   \multicolumn{1}{c|}{12.21}   \\ \cline{2-3} \cline{5-8} 
  &
     mul8s\_1KV8 &
  0-0-11-0--011 &
   &
  3.5 &
  12.00 &
 20470090
  &
   \multicolumn{1}{c|}{12.35}   \\ \cline{2-3} \cline{5-8} 
 &
     mul8s\_1KV8 &
  0-0-11-0--100 &
   &
  2.5 &
  12.15 &
 20470090
  &
   \multicolumn{1}{c|}{12.33}   \\ \cline{2-3} \cline{5-8} 
 & 
     mul8s\_1KV8 &
  0-0-00-0--001 &
   &
  0.0 &
  12.64 &

  20470090
 &
   \multicolumn{1}{c|}{12.43} \\ \hline
\end{tabular}}
\end{table*}

\begin{table*}[h]
\centering
\caption{Case study: the impact of full approximation on three different MLP architectures}
\label{case-study}
\resizebox{16.5cm}{!}{
\begin{tabular}{|c|c|c|c|c|c|c|c|}
\hline
\begin{tabular}[c]{@{}c@{}}Network\\ MNIST dataset\end{tabular}        & \begin{tabular}[c]{@{}c@{}}Exact network \\ accuracy (\%)\end{tabular} & \begin{tabular}[c]{@{}c@{}}Normalized \\ resource\\ utilization (\%)\\ {[}exact network{]}\end{tabular} & AxM  & \begin{tabular}[c]{@{}c@{}}Accuracy drop\\ (\%)\end{tabular} & \begin{tabular}[c]{@{}c@{}}Fault\\ vulnerability\end{tabular} & \begin{tabular}[c]{@{}c@{}}Normalized\\ latency\end{tabular} & \begin{tabular}[c]{@{}c@{}}Normalized \\resource\\ utilization (\%)\end{tabular} \\ \hline
\multirow{3}{*}{\begin{tabular}[c]{@{}c@{}}7-layer\\ MLP\end{tabular}} & \multirow{3}{*}{98.80}                                            & \multirow{3}{*}{100}                                                                                   & mul8s\_1KV8 & 0.2                                                          & 2.45                                                          & 1.00                                                         & 96                                                                            \\ \cline{4-8} 
                                                                       &                                                                   &                                                                                                      & mul8s\_1KV9 & 1.4                                                          & 1.03                                                          & 1.00                                                         & 90                                                                            \\ \cline{4-8} 
                                                                       &                                                                   &                                                                                                      & mul8s\_1KVP & 0.9                                                          & 1.33                                                          & 0.75                                                         & 76                                                                            \\ \hline
\multirow{3}{*}{\begin{tabular}[c]{@{}c@{}}5-layer\\ MLP\end{tabular}} & \multirow{3}{*}{86.30}                                            & \multirow{3}{*}{69}                                                                                & mul8s\_1KV8 & 0.0                                                          & 3.33                                                          & 1.00                                                         & 96                                                                            \\ \cline{4-8} 
                                                                       &                                                                   &                                                                                                      & mul8s\_1KV9 & 1.9                                                          & 2.12                                                          & 1.00                                                         & 89                                                                            \\ \cline{4-8} 
                                                                       &                                                                   &                                                                                                      & mul8s\_1KVP & 3.1                                                          & 3.84                                                          & 0.78                                                         & 76                                                                            \\ \hline
\multirow{3}{*}{\begin{tabular}[c]{@{}c@{}}3-layer\\ MLP\end{tabular}} & \multirow{3}{*}{80.40}                                            & \multirow{3}{*}{36}                                                                                & mul8s\_1KV8 & 0.4                                                          & 14.14                                                         & 1.00                                                         & 95                                                                            \\ \cline{4-8} 
                                                                       &                                                                   &                                                                                                      & mul8s\_1KV9 & 4.6                                                          & 7.62                                                          & 1.00                                                         & 88                                                                            \\ \cline{4-8} 
                                                                       &                                                                   &                                                                                                      & mul8s\_1KVP & 5.8                                                          & 9.54                                                          & 0.76                                                         & 74                                                                            \\ \hline

\end{tabular}}
\end{table*}
The proposed methodology is validated on three networks, i.e. a 3-layer MLP, LeNet-5 and AlexNet, trained on two representative datasets MNIST and Cifar-10. Each network is fully quantized down to 8-bit INT as a part of the preprocessing step of the methodology. The accuracy results for the quantized networks are reported in Table \ref{net}. Further, all possible combinations of approximate layers in the network are tested for selective approximation. For each experiment, three different multipliers reported in Table \ref{mult} are examined separately for efficiency to substitute the original exact multipliers. 

The fault injection procedure is performed for all different configurations, and the accuracy drop, due to approximation and fault injection, is profiled. Further, the HLS synthesis results of all configurations are generated, and the resource utilization in the number of FF, LUTs as well as the number of clock cycles required for processing one image for each network, are collected. A Pareto frontier for resource utilization and accuracy drop due to applying FI on different approximation configurations is plotted, and the results for LeNet-5 are reported in Fig. \ref{pareto}(a).

Fig. \ref{pareto}(b) shows the points on the Pareto frontier. The first column is the accuracy drop due to performing fault injection on that particular AxDNN configuration, the second column is resource utilization of the AxDNN in percentage, and finally, the last column is the selected approximate multiplier (AxM) and order of layers in ad-hoc (ones means that particular layer is approximated and dashes represent the non-computational layers like maxpooling). The coloured rows are some extreme and mid-range points of the Pareto chart.
The same experiment is repeated for MLP and AlexNet networks, and the results for some extreme and mid-range points of their pareto charts are presented in Table \ref{all-results}. 

It can be observed from this table that, generally, by approximating more layers, the latency and resource utilization are less. It is also noteworthy that the fault vulnerability of the network, which can be defined as the accuracy drop of the AxDNN due to applying FI, also becomes less. Fault vulnerability is opposite to fault resiliency and means the more the accuracy of an AxDNN drops due to applying FI, the more vulnerable the network is against faults. Generally, by increasing the level of approximation, the network shows better resiliency to faults. Still, there are several configurations that do not follow this trend and a tailored analysis using a framework such as DeepAxe is necessary for higher confidence. 

Fig. \ref{ax-accuracy} depicts the impact of different approximation units on the case-study DNNs' accuracy, resource utilization and fault vulnerability. For each network, three approximation units are chosen. For approximating the networks, the same configurations are picked to observe the impact of different AxM on the networks. Then all approximation units are applied, and the accuracy drop, fault vulnerability and resource utilization are reported. The correlation between the AxM error metrics reported in Table \ref{mult}, their area overhead, and the accuracy drop of the AxDNN impacted by AxMs lead us toward a conclusion that the network accuracy is generally impacted by a) the level of approximation and the configuration of the layers that are substituted by AxM; b) the error metrics of the AxM that is used as a substitution of ExC unit.

\subsection{Approximate multipliers case-study}
As a case study, three MLP networks with different architectures on the basis of a number of layers are selected. The base accuracy for each quantized network is 98.80\% for the network with 7 layers, 86.30\% for a network containing 5 layers and 80.40\% for 3-layer MLP network. The results for full approximation of the MLP networks with each case-study approximate multiplier (AxM) are reported in Table \ref{case-study}.

All the values in the table are normalized to the corresponding values of the ExC networks.

For the 7-layer MLP, it is shown that the multiplier mult8s\_KVP is the best option for full approximation, in the sense that the accuracy of the network drops only 0.9\%, and yet, latency and resource utilization of the network are better than for the other two multipliers. Therefore, based on the application of the network, if the designer can sacrifice the accuracy for 0.9\%, they can gain 25\% improvement in network latency and 24\% improvement in resource utilization of the implemented network on FPGA. 

The situation is different for the 5-layer MLP network. Based on the results of Table \ref{case-study}, the best multiplier can be mult8s\_KV9 since the accuracy does not drop dramatically and yet, it gains a better resiliency than the other two multipliers. Similarly, in the 3-layer MLP, the best candidate for full approximation of the network is mult8s\_KV9 multiplier since it shows the best resiliency with a little accuracy drop and still, provides 12\% improvement in resource utilization compared to the exact design. 

In summary, this case study shows the importance of exploring different AxMs for optimal implementation, i.e. not to compromise the accuracy of the network and, at the same time, to improve the network resiliency and hardware performance of the target design.

\section{Conclusion}

In this paper, we proposed a framework DeepAxe for design space exploration for FPGA-based implementation of DNNs by considering the trilateral impact of applying functional approximation on accuracy, reliability and hardware performance. 
The framework enables selective approximation of reliability-critical DNNs, providing a set of Pareto-optimal DNN implementation design space points for the target resource utilization requirements
The design flow starts with a pre-trained network in Keras, uses an innovative high-level synthesis environment DeepHLS and results in a set of Pareto-optimal design space points as a guide for the designer. The framework is demonstrated on a case study of custom and state-of-the-art DNNs and datasets.

\section{Acknowledgement}
This work was supported in part by the European Union through European Social Fund in the frames of the "Information and Communication Technologies (ICT) programme" (“ITA-IoIT” topic), by the Estonian Research Council grant PUT PRG1467, CRASHLESS“, by Estonian-French PARROT project "EnTrustED", by the Swedish Innovation Agency VINNOVA project “SafeDeep”, and Swedish Knowledge Foundation project “HERO”.
\bibliographystyle{IEEEtran}
\bibliography{ref}

\begin{thebibliography}{10}
\providecommand{\url}[1]{#1}
\csname url@samestyle\endcsname
\providecommand{\newblock}{\relax}
\providecommand{\bibinfo}[2]{#2}
\providecommand{\BIBentrySTDinterwordspacing}{\spaceskip=0pt\relax}
\providecommand{\BIBentryALTinterwordstretchfactor}{4}
\providecommand{\BIBentryALTinterwordspacing}{\spaceskip=\fontdimen2\font plus
\BIBentryALTinterwordstretchfactor\fontdimen3\font minus
  \fontdimen4\font\relax}
\providecommand{\BIBforeignlanguage}[2]{{%
\expandafter\ifx\csname l@#1\endcsname\relax
\typeout{** WARNING: IEEEtran.bst: No hyphenation pattern has been}%
\typeout{** loaded for the language `#1'. Using the pattern for}%
\typeout{** the default language instead.}%
\else
\language=\csname l@#1\endcsname
\fi
#2}}
\providecommand{\BIBdecl}{\relax}
\BIBdecl

\bibitem{taheri2022dnn}
M.~Taheri, ``Dnn hardware reliability assessment and enhancement,'' \emph{27th
  IEEE European Test Symposium (ETS).}, May 2022.

\bibitem{gholami2021survey}
A.~Gholami, S.~Kim, Z.~Dong, Z.~Yao, M.~W. Mahoney, and K.~Keutzer, ``A survey
  of quantization methods for efficient neural network inference,'' \emph{arXiv
  preprint arXiv:2103.13630}, 2021.

\bibitem{riazati2022autodeephls}
M.~Riazati, M.~Daneshtalab, M.~Sj{\"o}din, and B.~Lisper, ``Autodeephls: Deep
  neural network high-level synthesis using fixed-point precision,'' in
  \emph{2022 IEEE 4th International Conference on Artificial Intelligence
  Circuits and Systems (AICAS)}.\hskip 1em plus 0.5em minus 0.4em\relax IEEE,
  2022, pp. 122--125.

\bibitem{choudhary2022approximate}
P.~Choudhary, L.~Bhargava, V.~Singh, and A.~K. Suhag, ``Approximate computing:
  Evolutionary methods for functional approximation of digital circuits,''
  \emph{Materials Today: Proceedings}, 2022.

\bibitem{ibrahim2020soft}
Y.~Ibrahim, H.~Wang, J.~Liu, J.~Wei, L.~Chen, P.~Rech, K.~Adam, and G.~Guo,
  ``Soft errors in dnn accelerators: A comprehensive review,''
  \emph{Microelectronics Reliability}, vol. 115, p. 113969, 2020.

\bibitem{shafique2020robust}
M.~Shafique, M.~Naseer, T.~Theocharides, C.~Kyrkou, O.~Mutlu, L.~Orosa, and
  J.~Choi, ``Robust machine learning systems: Challenges, current trends,
  perspectives, and the road ahead,'' \emph{IEEE Design \& Test}, vol.~37,
  no.~2, pp. 30--57, 2020.

\bibitem{booth2022algorithm}
J.~D. Booth, ``Algorithm-based fault tolerance at scale,'' 2022.

\bibitem{bosio2019reliability}
A.~Bosio, P.~Bernardi, A.~Ruospo, and E.~Sanchez, ``A reliability analysis of a
  deep neural network,'' in \emph{2019 IEEE Latin American Test Symposium
  (LATS)}.\hskip 1em plus 0.5em minus 0.4em\relax IEEE, 2019, pp. 1--6.

\bibitem{cavagnero2022fault}
N.~Cavagnero, F.~D. Santos, M.~Ciccone, G.~Averta, T.~Tommasi, and P.~Rech,
  ``Fault-aware design and training to enhance dnns reliability with
  zero-overhead,'' \emph{arXiv preprint arXiv:2205.14420}, 2022.

\bibitem{siddique2021exploring}
A.~Siddique, K.~Basu, and K.~A. Hoque, ``Exploring fault-energy trade-offs in
  approximate dnn hardware accelerators,'' in \emph{2021 22nd International
  Symposium on Quality Electronic Design (ISQED)}.\hskip 1em plus 0.5em minus
  0.4em\relax IEEE, 2021, pp. 343--348.

\bibitem{2}
A.~Ghaffari and Y.~Savaria, ``Cnn2gate: Toward designing a general framework
  for implementation of convolutional neural networks on fpga,'' \emph{arXiv
  preprint arXiv:2004.04641}, 2020.

\bibitem{9294881}
M.~Riazati, M.~Daneshtalab, M.~Sjödin, and B.~Lisper, ``Deephls: A complete
  toolchain for automatic synthesis of deep neural networks to fpga,'' in
  \emph{2020 27th IEEE International Conference on Electronics, Circuits and
  Systems (ICECS)}, 2020, pp. 1--4.

\bibitem{venieris2018toolflows}
S.~I. Venieris, A.~Kouris, and C.-S. Bouganis, ``Toolflows for mapping
  convolutional neural networks on fpgas: A survey and future directions,''
  \emph{arXiv preprint arXiv:1803.05900}, 2018.

\bibitem{guo2019dl}
K.~Guo, S.~Zeng, J.~Yu, Y.~Wang, and H.~Yang, ``[dl] a survey of fpga-based
  neural network inference accelerators,'' \emph{ACM Transactions on
  Reconfigurable Technology and Systems (TRETS)}, vol.~12, no.~1, pp. 1--26,
  2019.

\bibitem{abdelouahab2018accelerating}
K.~Abdelouahab, M.~Pelcat, J.~Serot, and F.~Berry, ``Accelerating cnn inference
  on fpgas: A survey,'' \emph{arXiv preprint arXiv:1806.01683}, 2018.

\bibitem{molina2022high}
R.~S. Molina, V.~Gil-Costa, M.~L. Crespo, and G.~Ramponi, ``High-level
  synthesis hardware design for fpga-based accelerators: Models, methodologies,
  and frameworks,'' \emph{IEEE Access}, vol.~10, pp. 90\,429--90\,455, 2022.

\bibitem{FINN}
Y.~Umuroglu, N.~J. Fraser, G.~Gambardella, M.~Blott, P.~Leong, M.~Jahre, and
  K.~Vissers, ``Finn: A framework for fast, scalable binarized neural network
  inference,'' in \emph{Proceedings of the 2017 ACM/SIGDA international
  symposium on field-programmable gate arrays}, 2017, pp. 65--74.

\bibitem{1}
S.~I. Venieris and C.-S. Bouganis, ``fpgaconvnet: Mapping regular and irregular
  convolutional neural networks on fpgas,'' \emph{IEEE transactions on neural
  networks and learning systems}, vol.~30, no.~2, pp. 326--342, 2018.

\bibitem{ghaffari2020cnn2gate}
A.~Ghaffari and Y.~Savaria, ``Cnn2gate: Toward designing a general framework
  for implementation of convolutional neural networks on fpga,'' \emph{arXiv
  preprint arXiv:2004.04641}, 2020.

\bibitem{mousouliotis2020cnn}
P.~G. Mousouliotis and L.~P. Petrou, ``Cnn-grinder: from algorithmic to
  high-level synthesis descriptions of cnns for low-end-low-cost fpga socs,''
  \emph{Microprocessors and Microsystems}, vol.~73, p. 102990, 2020.

\bibitem{stone2010opencl}
J.~E. Stone, D.~Gohara, and G.~Shi, ``Opencl: A parallel programming standard
  for heterogeneous computing systems,'' \emph{Computing in science \&
  engineering}, vol.~12, no.~3, p.~66, 2010.

\bibitem{guo2016angel}
K.~Guo, L.~Sui, J.~Qiu, S.~Yao, S.~Han, Y.~Wang, and H.~Yang, ``Angel-eye: A
  complete design flow for mapping cnn onto customized hardware,'' in
  \emph{2016 IEEE Computer Society Annual Symposium on VLSI (ISVLSI)}.\hskip
  1em plus 0.5em minus 0.4em\relax IEEE, 2016, pp. 24--29.

\bibitem{sharma2016high}
H.~Sharma, J.~Park, D.~Mahajan, E.~Amaro, J.~K. Kim, C.~Shao, A.~Mishra, and
  H.~Esmaeilzadeh, ``From high-level deep neural models to fpgas,'' in
  \emph{2016 49th Annual IEEE/ACM International Symposium on Microarchitecture
  (MICRO)}.\hskip 1em plus 0.5em minus 0.4em\relax IEEE, 2016, pp. 1--12.

\bibitem{jouppi2017datacenter}
N.~P. Jouppi, C.~Young, N.~Patil, D.~Patterson, G.~Agrawal, R.~Bajwa, S.~Bates,
  S.~Bhatia, N.~Boden, A.~Borchers \emph{et~al.}, ``In-datacenter performance
  analysis of a tensor processing unit,'' in \emph{Proceedings of the 44th
  annual international symposium on computer architecture}, 2017, pp. 1--12.

\bibitem{sarwar2018energy}
S.~S. Sarwar, S.~Venkataramani, A.~Ankit, A.~Raghunathan, and K.~Roy,
  ``Energy-efficient neural computing with approximate multipliers,'' \emph{ACM
  Journal on Emerging Technologies in Computing Systems (JETC)}, vol.~14,
  no.~2, pp. 1--23, 2018.

\bibitem{hong2019terminal}
S.~Hong, P.~Frigo, Y.~Kaya, C.~Giuffrida, and T.~Dumitraș, ``Terminal brain
  damage: Exposing the graceless degradation in deep neural networks under
  hardware fault attacks,'' in \emph{28th USENIX Security Symposium (USENIX
  Security 19)}, 2019, pp. 497--514.

\bibitem{kundu2020high}
S.~Kundu, A.~Soyyi{\u{g}}it, K.~A. Hoque, and K.~Basu, ``High-level modeling of
  manufacturing faults in deep neural network accelerators,'' in \emph{2020
  IEEE 26th International Symposium on On-Line Testing and Robust System Design
  (IOLTS)}.\hskip 1em plus 0.5em minus 0.4em\relax IEEE, 2020, pp. 1--4.

\bibitem{li2017understanding}
G.~Li, S.~K.~S. Hari, M.~Sullivan, T.~Tsai, K.~Pattabiraman, J.~Emer, and S.~W.
  Keckler, ``Understanding error propagation in deep learning neural network
  (dnn) accelerators and applications,'' in \emph{Proceedings of the
  International Conference for High Performance Computing, Networking, Storage
  and Analysis}, 2017, pp. 1--12.

\bibitem{zhang2018analyzing}
J.~J. Zhang, T.~Gu, K.~Basu, and S.~Garg, ``Analyzing and mitigating the impact
  of permanent faults on a systolic array based neural network accelerator,''
  in \emph{2018 IEEE 36th VLSI Test Symposium (VTS)}.\hskip 1em plus 0.5em
  minus 0.4em\relax IEEE, 2018, pp. 1--6.

\bibitem{deepvigor}
M.~H. Ahmadilivani, M.~Taheri, J.~Raik, M.~Daneshtalab, and M.~Jenihhin,
  ``Deepvigor: Vulnerability value ranges and factors for dnns reliability
  assessment,'' in \emph{28th IEEE European Test Symposium}.\hskip 1em plus
  0.5em minus 0.4em\relax In press, 2023.

\bibitem{bushnell2004essentials}
M.~Bushnell and V.~Agrawal, \emph{Essentials of electronic testing for digital,
  memory and mixed-signal VLSI circuits}.\hskip 1em plus 0.5em minus
  0.4em\relax Springer Science \& Business Media, 2004, vol.~17.

\bibitem{david2021tensorflow}
R.~David, J.~Duke, A.~Jain, V.~Janapa~Reddi, N.~Jeffries, J.~Li, N.~Kreeger,
  I.~Nappier, M.~Natraj, T.~Wang \emph{et~al.}, ``Tensorflow lite micro:
  Embedded machine learning for tinyml systems,'' \emph{Proceedings of Machine
  Learning and Systems}, vol.~3, pp. 800--811, 2021.

\bibitem{evoapprox16}
V.~Mrazek, R.~Hrbacek, Z.~Vasicek, and L.~Sekanina, ``Evoapprox8b: Library of
  approximate adders and multipliers for circuit design and benchmarking of
  approximation methods,'' in \emph{Design, Automation Test in Europe
  Conference Exhibition (DATE), 2017}, March 2017, pp. 258--261.

\bibitem{leveugle2009statistical}
R.~Leveugle, A.~Calvez, P.~Maistri, and P.~Vanhauwaert, ``Statistical fault
  injection: Quantified error and confidence,'' in \emph{2009 Design,
  Automation \& Test in Europe Conference \& Exhibition}.\hskip 1em plus 0.5em
  minus 0.4em\relax IEEE, 2009, pp. 502--506.

\end{thebibliography}

\end{document}